\documentclass{article}

\usepackage{times}
\usepackage{graphicx} 
\usepackage{subfigure}
\usepackage{amsmath}
\usepackage{amsthm}
\usepackage{bbm}
\usepackage{tikz}
\newenvironment{varalgorithm}[1]
  {\algorithm\renewcommand{\thealgorithm}{#1}}
  {\endalgorithm}

\usepackage{natbib}

\usepackage{algorithm}
\usepackage{algorithmic}

\usepackage{hyperref}

\usepackage[accepted]{icml2015}

\newtheorem{theorem}{Theorem}

\begin{document}

\twocolumn[

\icmltitle{Learning Scale-Free Networks by Dynamic Node-Specific Degree Prior}

\icmlauthor{Qingming Tang\footnotemark[1]}{qmtang@ttic.edu}
\icmladdress{Toyota Technological Institute at Chicago,
            6045 S. Kenwood Ave., Chicago, Illinois 60637, USA}
\icmlauthor{Siqi Sun\footnotemark[1]}{siqi@ttic.edu}
\icmladdress{Toyota Technological Institute at Chicago,
            6045 S. Kenwood Ave., Chicago, Illinois 60637, USA}
\icmlauthor{Jinbo Xu}{jinbo.xu@gmail.com}
\icmladdress{Toyota Technological Institute at Chicago,
            6045 S. Kenwood Ave., Chicago, Illinois 60637, USA}

\icmlkeywords{scale-free, graphical model, structured prediction}

\vskip 0.3in
]

\begin{abstract}
Learning network structure underlying data is an important problem in machine learning. This paper presents a novel degree prior to study the inference of scale-free networks, which are widely used to model social and biological networks. In particular, this paper formulates scale-free network inference using Gaussian Graphical model (GGM) regularized by a node degree prior. Our degree prior not only promotes a desirable global degree distribution, but also exploits the estimated degree of an individual node and the relative strength of all the edges of a single node. To fulfill this, this paper proposes a ranking-based method to dynamically estimate the degree of a node, which makes the resultant optimization problem challenging to solve. To deal with this, this paper presents a novel ADMM (alternating direction method of multipliers) procedure. Our experimental results on both synthetic and real data show that our prior not only yields a scale-free network, but also produces many more correctly predicted edges than existing scale-free inducing prior, hub-inducing prior and the $l_1$ norm.
\end{abstract}
\footnotetext[1]{Equal Contribution.}
\section{Introduction}
Graphical models are widely used to describe the relationship between variables (or features).
Estimating the structure of an undirected graphical model from a dataset has been extensively studied \cite{meinshausen2006high, yuan2007model, friedman2008sparse,banerjee2008model, wainwright2006high}. Gaussian Graphical Models (GGMs) are widely used to model the data and $l_1$ penalty is used to yield a sparse graph structure. GGMs assume that the observed data follows a multivariate Gaussian distribution with a covariance matrix. When GGMs are used, the network structure is encoded in the zero pattern of the precision matrix. Accordingly, structure learning can be formulated as minimizing the negative log-likelihood (NLL) with an $l_1$ penalty.
However, the widely-used $l_1$ penalty assumes that any pair of variables is equally likely to form an edge.
This is not suitable for many real-world networks such as gene networks, protein-protein interaction networks and social networks, which are scale-free and contain a small percentage of hub nodes. \par

A scale-free network \cite{barabasi2003scale} has node degree following the power-law distribution.
In particular, a scale-free network may contain a few hub nodes, whose degrees are much larger than the others. That is, a hub node more likely forms an edge than the others. In real-world applications, a hub node is usually functionally important. For example, in a gene network, a gene playing functions in many biological processes \cite{zhang2005general,goh2007human} tends to be a hub.
A few methods have been proposed to infer scale-free networks by using a reweighed $l_1$ norm. For example, \cite{peng2009partial} proposed a joint regression sparse method to learn scale-free networks by setting the penalty proportional to the estimated degrees. \cite{candes2008enhancing} proposed a method that iteratively reweighs the penalty in the $l_1$ norm based on the inverse of the previous estimation of precision matrix. This method can suppress the large bias in the $l_1$ norm when the magnitude of the non-zero entries vary a lot and is also closer to the $l_0$ norm. Some recent papers follow the idea of node-based learning using group lasso \cite{friedman2010note,meier2008group, tan2014learning,friedman2010regularization,mohan2014node}. Since group lasso penalty promotes similar patterns among variables in the same group, it tends to strengthen the signal of hub nodes and suppress that of non-hub nodes. As such, group lasso may only produce edges adjacent to nodes with a large degree (i.e., hubs). \par

 In order to capture the scale-free property, \cite{liu2011learning} minimizes the negative log-likelihood with a scale-free prior, which approximates the degree of each variable by the $l_1$ norm. However, the objective function is non-convex and thus, is hard to optimize. \cite{defazio2012convex} approximates the global node degree distribution by the submodular convex envelope of the node degree function. The convex envelope is the Lovasz extension \cite{lovasz1983submodular,bach2010structured} of the sum of logarithm of node degree. These methods consider only the global node degree distribution, but not the degree of an individual node. However, the approximation function used to model this global degree distribution may not induce a power-law distribution, because there are many possible distributions other than power-law that optimizing the approximation function. See \textbf{Theorem} \ref{global_problem} for details. \par

To further improve scale-free network inference, this paper introduces a novel node-degree prior, which not only promotes a desirable global node degree distribution, but also exploits the estimated degree of an individual node and the relative strengths of all the edges adjacent to the same node.
To fulfill this, we use node and edge ranking to dynamically estimate the degree of an individual node and the relative strength of one potential edge.
As dynamic ranking is used in the prior, the objective function is very challenging to optimize.
This paper presents a novel ADMM (alternating direction method of multipliers) \cite{boyd2011distributed, fukushima1992application} method for this.

\section{Model}

\subsection{Notation \& Preliminaries}
Let $G=(V,E)$ denote a graph, where $V$ is the vertex set ($p=|V|$) and $E$ the edge set. We also use $E$ to denote the adjacency matrix of $G$, i.e., $E_{ij} = 1$ if and only if $(i,j)\in E$. In this paper we assume that the variable set $V$ has a Gaussian distribution with a precision matrix $X$. Then, the graph structure is encoded in the zero pattern of the estimated $X$, i.e. the edge formed by $X$ is $E(X) = \{(u,v),  X_{uv}\ne 0\}$. Let $F(X)$ denote the objective function. For GGMs, $F(X) = tr(X S) - \log \det(X)$, where $S$ is the empirical covariance matrix. To induce a sparse graph, a $l_1$ penalty is applied to obtain the following objective function.
\begin{align}
\hat{X} = \arg\min_X F(X) + \lambda \| X \|_1
\end{align}

Here we define two ranking functions. For a given $i$, let $(.)$ denote the permutation of $(1,2,...,p-1)$ such that $|X_{i,(1)}| \geq |X_{i,(2)}| \geq ... \geq |X_{i,(p-1)}|$ where $(X_{i,(1)},X_{i,(2)},...,X_{i,(p-1)})$ is a shuffle of a row vector $X_i$ excluding the element $X_{i,i}$.
Let $M$ be a $p-1$ dimension positive vector and $V$ a $p$ dimension vector.  We define $X_i \circ M= \sum_{u=1}^{p-1}{|X_{i,(u)}|M_u}$.
Let $[X,M,V]$ be a permutation of $(1,2,...,p)$ such that $X_{[1,X,M,V]}\circ M+V_1 \geq X_{[2,X,M,V]}\circ M + V_2 \geq ... \geq X_{[p,X,M,V]}\circ M+V_p$, where $[j,X,M,V]$ denote the $j^{th}$ element of $[X,M,V]$. When $V$ is a zero vector, we also write $[X,M,V]$ as $[X,M]$ and $[j,X,M,V]$ as $[j,X,M]$, respectively.\par

To promote a sparse graph, a natural way is to use a $l_1$ penalty to regulate $F(X)$. However, $l_1$ norm cannot induce a scale-free network. This is because that a scale-free network has node degree following a power-law distribution, i.e., $P(d)\propto d^{-\gamma}$ where $d$ is the degree and $\gamma$ is a constant ranging from 2 to 3, rather than a uniform distribution. A simple prior for scale-free networks is the sum of the logarithm of node degree as follows.
\begin{align}
\lambda \sum_{v=1}^{p} \gamma \log(d_v + 1),
\label{ori_obj}
\end{align}
where $d_v = \sum_{u=1}^p I[X_{vu}\ne 0,u \neq v]$ and $I$ is the indicator function. The constant 1 is added to handle the situation when $d_v=0$. The $l_0$ norm in (\ref{ori_obj}) is non-differentiable and thus, hard to optimize. One way to resolve this problem is to approximate $d_v$ by the $l_1$ norm of $X_v$ as shown in \cite{liu2011learning}. Since the logarithm of $l_1$ approximation is non-convex, a convex envelope of (\ref{ori_obj}) based on submodular function and Lovasz extension \cite{lovasz1983submodular} is introduced recently in the papers \cite{bach2010structured, defazio2012convex} as follows.
\begin{align}
\sum_{v=1}^p{\sum_{u=1}^{p-1} {|X_{v,(u)}| \Big(\log(u+1) - \log (u) \Big)}}.
\label{lov_ori}
\end{align}
Let $h(u) =  (\log(u+1)-\log(u))$. The convex envelope of Eq. \ref{ori_obj} can be written as
\begin{align}
\sum_{v=1}^p{\sum_{u=1}^{p-1} {h(u)|X_{v,(u)}|}}.
\end{align}

\subsection{Node-Specific Degree Prior}
Although Eq. \ref{ori_obj} and its approximations can be used to induce a scale-free graph, they still have some issues, as explained in \textbf{Theorem} \ref{global_problem}. 
\begin{theorem}
\label{global_problem}
Let $S(G)$ denote the sum of logarithm of node degree of a graph $G$. For any graph $G$ satisfying the scale-free property, there exists another graph $G'$ with the same number of edges but not satisfying the scale-free property such that $S(G')<S(G)$.
\end{theorem}
Since we would like to use (\ref{ori_obj}) as a regularizer, a smaller value of (\ref{ori_obj}), denoted as $S(G)$, is always favorable. Here we briefly prove \textbf{Theorem} \ref{global_problem}. Since $G$ is scale-free, a large percentage of its nodes have small degrees. It is reasonable to assume that $G$ has two nodes $u'$ and $v'$ with the same degree. Let $a$ denote their degree. Without loss of generality, there exists a node $x$ connecting to $u'$ but not to $v'$. We construct a graph $G^{(1)}$ from $G$ by connecting $x$ to $v'$ and disconnecting $x$ from $u'$. Since $\log$ is a concave function, we have
\begin{align}
S(G^{(1)})&=S(G)-2\log(a+1) + \log(a) + \log(a+2) \nonumber \\
&< S(G)
\label{reason}
\end{align}
If there are two nodes with the same degree in $G^{(i)}$, we can construct $G^{(i+1)}$ from $G^{(i)}$ such that $S(G^{(+1)})<S(G^{(i)})<....<S(G^{(1)})<S(G)$. By this procedure, we may finally obtain a non-scale-free graph  with a smaller $S(G)$ value. 
So, \textbf{Theorem} \ref{global_problem} is proved. \par

Since \textbf{Theorem} \ref{global_problem} shows that a global description of scale-free property is not the ideal choice for inducing a scale-free graph, we propose a new prior to describe the degree of each individual node in the target graph.  \par
Intuitively, given a node $u$, if we rank all the potential edges adjacent to $u$ in descending order by $|X_{u,v}|$ where $v \in V-{u}$, then $(u,v)$ with a higher rank is more likely to be a true edge than those with a lower rank and should receive less penalty. To account for this intuition, we introduce a non-decreasing positive sequence $H$ and the following prior for node $u$
\begin{equation}
H\circ X_u = \sum_{v=1}^{p-1}H_v |X_{u,(v)}|.
\label{single_node_penalty}
\end{equation}
Here $(v)$ represents the node ranked in the $v^{th}$ position. When all the elements in $H$ are same, this prior  simply is $l_1$ norm. In our work, we use a moderately increasing positive sequence $H$ (e.g. $H_v={\log(v+1)}^{\alpha}$ for $v=1,2,...,p-1$), such that a pair $(u,v)$ with a larger estimated $|X_{u,v}|$ has a smaller penalty. On the other hand, we do not want to penalize all the edges very differently since the penalty is based upon only rough estimation of the edge strength.  \par

Let $d_u$ denote the degree of node $u$, we propose the following prior based on Eq. \ref{single_node_penalty}.
\begin{equation}
\sum_{u=1}^{p}{\frac{H\circ X_u}{H_{d_u}}}
\label{node_specific}
\end{equation}
Comparing to the $l_1$ norm, our prior in Eq.\ref{node_specific} is non-uniform since each edge has a different penalty, depending on its ranking and the estimated node degrees. This prior has the following two properties.
\begin{itemize}
	\item When $v \leq d_u$, the penalty for $X_{u,(v)}$ is less than or equal to $1$; otherwise, the penalty is larger than 1. 
	\item If node $i$ has a higher degree than node $j$, then $X_{i,(u)}$ has a smaller penalty than $X_{j,(u)}$.
\end{itemize}
The first property implies that Eq.(\ref{node_specific}) favors a graph following a given degree distribution. Suppose that in the true graph, node (or variable) $i$ ($i=1,2,...,p$) has degree $d_i$. Let $E^{(1)}$ and $E^{(2)}$ denote two edge sets of the same size. If $E^{(1)}$ has the degree distribution $(d_1,d_2,...,d_p)$ but $E^{(2)}$ does not, then we can prove the following result (see Supplementary for proof).
\begin{align}
\sum_{u=1}^{p}{ \frac{H\circ E^{(1)}_u}{H_{d_u}} } \leq \sum_{u=1}^{p}{ \frac{H\circ E^{(2)}_u}{H_{d_u}} }
\end{align}
The second property implies that Eq.(\ref{node_specific}) is actually robust. Even if we cannot exactly estimate the node degree  $(d_1,d_2,...,d_p)$, Eq.(\ref{node_specific}) may still work well as long as we can accurately rank nodes by their degrees. 
\subsection{Dynamic Node-Specific Degree Prior}
We have introduced a prior (\ref{node_specific}) that favors a graph with a given degree distribution. Now the question is how to use (\ref{node_specific}) to produce a scale-free graph without knowing the exact node degree. \par
Let $\#_k$ denote the expected number of nodes with degree $k$. We can estimate $\#_k$ from the prior degree distribution $p_0(d) \propto d^{-\gamma}$ when $\gamma$ is given. Supposing that all the nodes are ranked in descending order by their degree, we use $\tau_i$ to denote the estimated degree of the $i^{th}$ ranked node based on the power-law distribution, such that $\tau_1 \ge \tau_2 \ge \dots \ge \tau_p$. Further, if we know the desired number of predicted edges, say we are told to output $a$ edges, then the expected degree $\tau'_v$ of node $v$ is assumed to be proportional to $a\times \frac{\tau_v}{\sum_{i=1}^{p}\tau_i}$. In the following content, we would just use $\tau_v$ to denote the expected degree of node $v$. \par 
Now the question is how to rank nodes by their degrees? Although the exact degree $d_v$ of node $v$ is not available, we can approximate $\log(d_v+1)$ by Lovasz Extenion \cite{defazio2012convex,bach2010structured,lovasz1983submodular}, i.e., $\sum_{u=1}^{p-1} {|X_{v,(u)}| \Big(\log(u+1) - \log (u) \Big)}$ (see Eq. \ref{ori_obj}). That is, we can rank nodes by their Lovasz Extension. Note that although we use Lovasz Extension in our implementation, other approximations such as $l_1$ norm can also be used to rank nodes without losing accuracy. \par
We define our dynamic node-specific prior as follows.
\begin{align}
\label{prior}
\Omega(X) = \sum_{v=1}^p{\frac{X_{[v,X,h]}\circ H}{H_{\tau_v}}}
\end{align}
Note that $h(u)=\log(u+1) - \log (u)$ for $1\leq u \leq p$ and $[X,h]$ defines the ranking based on Lovasz Extension. The $i$-th ranked node is assigned a node penalty $\frac{1}{H_{\tau_i}}$, denoted as $g(i)$. Note that the ranking of nodes by their degrees is not static. Instead, it is determined dynamically by our optimization algorithm. Whenever there is a new estimation of $X$, the node and edge ranking may be changed. That is, our node-specific degree prior is dynamic instead of static.

\section{Optimization}
With prior defined in (\ref{prior}), we have the following objective function:
\begin{align}
\label{final_obj}
F(x) & + \beta \Omega(X)
\end{align}
Here $\beta$ is used to control sparsity level. The challenge of minimizing Eq.(\ref{final_obj}) lies in the fact that we do not know the ranking of both nodes and edges in advance. Instead we have to determine the ranking dynamically. We use an ADMM algorithm to solve it by introducing dual variables $Y$ and $U$ as follows.
\begin{align}
\label{update_x}
X^{l+1} =& \arg \min_X F(X) + \frac{\rho}{2} \| X - Y^l + U^l \|^2_F \\
\label{update_y}
Y^{l+1} = & \arg \min_Y \beta \Omega(Y) + \frac{\rho}{2} \| X^{l+1} - Y + U^l \|^2_F \\
\label{update_u}
U^{l+1} = & U^l + X^{l+1} - Y^{l+1}
\end{align}

We can use a first-order method such as gradient descent to optimize the first subproblem since $F(X)$ is convex. In this paper, we assume that $F(x)$ is a Gaussian likelihood, in which case (\ref{update_x}) can be solved by eigen-decomposition. Here we describe a novel algorithm for (\ref{update_y}). \par

Let $A = X^{l+1}+U^{l}$ and $\lambda = \frac{\beta}{\rho}$. Minimizing (\ref{update_y}) is equivalent to
\begin{align}
\label{before_dual}
\min_Y & \frac{1}{2} \| Y - A \|^2_F + \lambda \Omega(Y),
\end{align}
which can be relaxed as
\begin{align}
\min_{Y} & \frac{1}{2} \| Y - A \|^2_F  + \lambda \sum_{v=1}^p g(v)  Y_{[v]} \circ H   \nonumber \\
\label{dual_original}
& s.t.  \quad g([v])=g([v, Y, h]) \quad 1\leq v \leq p.
\end{align}

Here, we simply use $[\cdot]$ to denote a permutation of $\{1,2,...,p-1\}$, and use $[v]$ to denote the $v^{th}$ element of this permutation. The reason we introduce (\ref{dual_original}) is, given $Y$ and without the constraint of (\ref{dual_original}), the optimal $[.]$ is $[Y,H]$. Adding the constraint $g([v])=g([v, Y, h])$, (\ref{dual_original}) can be relaxed by solving the following problem until $g([v,Y,H,\delta])=g([v, Y, h]) \quad 1\leq v \leq p$. 
\begin{align}
\min_{Y,\delta} & \frac{1}{2} \| Y - A \|^2_F  + \lambda \sum_{v=1}^p g(v) \{ Y_{[v,Y,H,\delta]} \circ H \}  \nonumber \\
\label{dual_decomp}
\end{align}
Here $\delta$ is the dual vector and can be updated by $\delta(v)=\mu \cdot (g([v,Y,H,\delta])-g([v, Y, h])$ for $1\leq v \leq p$, where $\mu$ is the step size. Actually, as only a small percentage of nodes have large degrees, we may speed up by using the condition $g([v,Y,H,\delta])=g([v, Y, h])$ for $\quad 1\leq v \leq k$ where $k$ is much smaller than $p$. That is, instead of ranking all the nodes, we just need to select top $k$ of them, which can be done much faster.
We now propose \textbf{Algorithm 1} to solve (\ref{dual_decomp}), which in spirit is a  dual decomposition \cite{sontag2011introduction} algorithm.

\begin{varalgorithm}{1}
\caption{Update of node ranking}
\begin{algorithmic}[1]
\STATE  Randomly generate $Y^0$. Set $t=0, \delta^0=0$ and compute $[Y^0, H, \delta^0]$
\WHILE {TRUE}
\STATE  $Y^{t+1} = \arg \min_Y \frac{1}{2} \| Y - A \|^2_F$
\STATE \ \ \ \ \ \ \ \ $+ \lambda \sum_v g(v) Y_{[v,Y^{t}, H, \delta^{t}]} \circ H $
\IF {$[Y^{t+1}, H, \delta^{t}] = [Y^{t}, H, \delta^{t}] = [Y^{t+1}, h]$}
\STATE break
\ELSE
\STATE $\delta^{t+1} = \delta^{t} + \mu ( g([Y^{t+1}, H, \delta^{t}])-g([Y^{t+1}, h]))$
\ENDIF
\STATE $ t=t+1$
\ENDWHILE
\label{node_rank_algo}
\end{algorithmic}
\end{varalgorithm}

\begin{theorem}
The output $Y'$ of algorithm \ref{node_rank_algo} satisfies the following condition.
\begin{align}
Y' = \arg\min_{Y}\{ \frac{1}{2} \| Y - A \|^2_F  + \lambda \sum_{v=1}^p g(v)  Y_{[v, Y', h]} \circ H  \}
\end{align}
\label{local_condition}
\end{theorem}
See supplementary for the proof of \textbf{Theorem} \ref{local_condition}. \par
Solving line 3-4 in \textbf{Algorithm} \ref{node_rank_algo} is not trivial. 
We can reformulate it as follows.
\begin{align}
\label{sub}
\min_{Y} \sum_{v=1}^p \{ \frac{1}{2} \| Y_v - A_v \|^2_F + Y_v \circ H'(v)\}.
\end{align}
Here $H'(v)$ is a $p-1$ dimension vector and $H'_u(v) = \lambda g(v) H_u$ for $1\leq u \leq p-1$. Since $Y$ is symmetric, (\ref{sub}) can be reformulated as follows.
\begin{align}
\label{reformat_sub}
\min_{Y} \sum_{v=1}^p &\{ \frac{1}{2} \| Y_v - A_v \|^2_F + Y_v \circ H'(v)\} \\ \nonumber
s.t. &\quad Y = Y^{T}.
\end{align}

Problem (\ref{reformat_sub}) can be solved iteratively using dual decomposition by introducing the Lagrangian term $tr(\sigma(Y-Y^{T}))$, where $\sigma$ is a $p$ by $p$ matrix which would be updated by $\sigma =\sigma+ \mu(Y-Y^{T})$. Notice that $tr(\sigma(Y-Y^{T}))=tr((\sigma-\sigma^T)Y)$, (\ref{reformat_sub}) can be decomposed into $p$ independent sub-problems as follows.
\begin{align}
\label{subsub}
\min_{Y_v}{\frac{1}{2}\|Y_v-(A + \sigma^T - \sigma)_v\|_F^2+Y_v \circ H'(v)}
\end{align}
Let $B=A + \sigma^T - \sigma$. Obviously $Y_{v,v}=B_{v,v}$ holds, so we do not consider $Y_{v,v}$ in the remaining section. Let  $\hat{Y}_{v}$=$\{\hat{Y}_{v,(1)},\hat{Y}_{v,(2)},...,\hat{Y}_{v,(p-1)}\}$ be a feasible solution of (\ref{subsub}). We define the cluster structure of $\hat{Y}_{v}$ as follows.\par

\textbf{Definition 3} Let $\{\hat{Y}_{v,(1)},\hat{Y}_{v,(2)},...,\hat{Y}_{v,(p-1)}\}$ be a ranked feasible solution. Supposing that $|\hat{Y}_{v,(1)}|=|\hat{Y}_{v,(2)}|=...=|\hat{Y}_{v,(t)}|>|\hat{Y}_{v,(t+1)}|$, we say $\{\hat{Y}_{v,(1)},\hat{Y}_{v,(2)},...,\hat{Y}_{v,(t)}\}$ form cluster 1 and denote it as $C|_{\hat{Y}_v}(1)$. Similarly, we can define $C|_{\hat{Y}_v}(2)$, $C|_{\hat{Y}_v}(3)$ and so on. Assume that $\{\hat{Y}_{v,(1)},\hat{Y}_{v,(2)},...,\hat{Y}_{v,(p-1)}\}$ is clustered into $T(\hat{Y}_v)$ groups. Let $\big|C|_{\hat{Y}_v}(k)\big|$ denote the size of $C|_{\hat{Y}_v}(k)$ and $y|_{\hat{Y}_v}(k)$ the absolute value of its element.

Assuming that $Y^{*}_v$ is the optimal solution of (\ref{subsub}),
by Definition 3,  for $1 \leq k \leq T(Y^{*}_v)$,  $Y^{*}_v$ has the following property. 
\begin{equation}
y|_{Y^{*}_v}(k) = \max\{0,\frac{\sum_{i \in {C|_{Y^{*}_v}(k)}}\{|B_{v,(i)}|-H'_i(v)\}}{\big|C|_{Y^{*}_v}(k) \big|}\}
\label{opt_property}
\end{equation}
See \cite{defazio2012convex} and our supplementary material for detailed proof of (\ref{opt_property}). Based on (\ref{opt_property}), we propose a novel dynamic programming algorithm to solve (\ref{subsub}), which can be reduced to the problem of finding the constrained optimal partition of $\{B_{v,(1)}, B_{v,(2)},...,B_{v,(p-1)}\}$. \par
Let $Y_{v,1:t}^*=\{C|_{Y_{v,1:t}^*}(1),C|_{Y_{v,1:t}^*}(2),...,C|_{Y_{v,1:t}^*}(m=T(Y_{v,1:t}^*))\}$ be the optimal partition for $\{B_{v,(1)}, B_{v,(2)},...,B_{v,(t)}\}$. Let $C|_{Y_{v,1:t}^*}(m+1)=\{|B_{v,(t+1)}|-H_{t+1}'(v)\}$ and $C|_{Y_{v,1:t}^*}(0)=\{ \infty \}$. Then we have the following theorem.
\begin{theorem}
$Y_{v,1:t+1}^*=\{C|_{Y_{v,1:t}^*}(1),...,C|_{Y_{v,1:t}^*}(k-1),C_k\}$, where $C_k$ is a set with $\Sigma_{s=k}^{m+1}{|C|_{Y_{v,1:t}^*}(s)}|$ elements which are equal to $y_k$.
\begin{align}
y_k = \max \{0,\frac{\sum_{i \in {\bigcup_{s=k}^{m+1}{C|_{Y_{v,1:t}^*}(s)}}}\{|B_{v,(i)}|-H'_i(v)\}}{\Sigma_{s=k}^{m+1}{|C|_{Y_{v,1:t}^*}(s)}|} \}
\end{align}
and $k$ is the largest value such that $y|_{Y_{v,1:t}^*}(k-1)>y_k$.
\label{problem_structure}
\end{theorem}
\textbf{Theorem} \ref{problem_structure} clearly shows that this problem satisfies the optimal substructure property and thus, can be solved by dynamic programming. A $O(p\log(p))$ algorithm (\textbf{Algorithm} \ref{node_ranking}) to solve (\ref{subsub}) is proposed.
See supplementary material for the proof of substructure property and the correctness of the $O(p\log(p))$ algorithm. In \textbf{Algorithm 2}, $Rep(x, p)$ duplicates $x$ by $p$ times.

\begin{varalgorithm}{2}
\caption{Edge rank updating}
\begin{algorithmic}[1]
\STATE Input: $B_v$ and $H'(v)$
\STATE Output: $Y^{*}_v$
\STATE Sort $B_v$, get $\{B_{v,(1)}, B_{v,(2)},...,B_{v,(p-1)}\}$
\STATE Initialize $t=0$, $sum(0)=0$ and $Y^{*}_{1..0}=\emptyset$
\STATE $t=t+1$
\WHILE {$t < p$}
\STATE $sum(t)=sum(t-1)+|B_{v,(t)}|-H'_t(v)$
\STATE $m=T(Y^{*}_{1..t-1})$, $C|_{Y_{v,1:t}^*}(m+1)=\{|B_{v,(t+1)}|-H_{t+1}'(v)\}$ and $C|_{Y_{v,1:t}^*}(0)=\{ \infty \}$
\STATE Binary search to find out the largest index $b$ among $1,...,m+1$, such that $y|_{Y_{v,1:t}^*}(b-1)>\max \{0,\frac{\sum_{i \in {\bigcup_{s=b}^{m+1}{C|_{Y_{v,1:t}^*}(s)}}}\{|B_{v,(i)}|-H'_i(v)\}}{\Sigma_{s=b}^{m+1}{|C|_{Y_{v,1:t}^*}(s)}|} \}$
\STATE $S = |\cup_{s=1}^{b-1}{C|_{Y^{*}_{1..t-1}}(s)}|$
\begin{align*}
\text{Set } & C|_{Y^{*}_{1..t}}(b) = \\
& \{Rep(max\{\frac{sum(i)-sum(S)}{t-S},0\},t-S)
\end{align*}
\STATE $Y^{*}_{1..t} = \cup_{s=1}^{b-1}{C|_{Y^{*}_{1..t-1}}(s)} + C|_{Y^{*}_{1..t}}(b)$
\ENDWHILE
\end{algorithmic}
\label{node_ranking}
\end{varalgorithm}

\section{Related Work \& Hyper Parameters}
To introduce scale-free prior ($p(d) \propto d^{-\alpha}$),  \cite{liu2011learning} proposed to approximate the degree of node $i$ by $\|X_{-i}\|_1=\sum_{j\ne i} |X_{i,j}|$ and then use the following objective function.
\begin{align}
F(X) + \alpha \sum_{i=1}^p \log( \|X_{-i}\|_1 + \epsilon_i)+ \beta\sum_{i=1}^p {|X_{i,i}|},
\end{align}
where $\epsilon_i$ is the parameter for each node to smooth out the scale-free distribution. Without prior knowledge, it is not easy to determine the value of $\epsilon_i$. Note that the above objective function is not convex even when $F(x)$ is convex because of the log-sum function involved. The objective is optimized by reweighing the penalty for each $X_{i,j}$ at each step, and the method (denoted as RW) is guaranteed to converge to a local optimal. The parameter $\epsilon$ is set as diagonal of estimated $X$ in previous iteration, and $\beta$ as $2\frac{\alpha}{\epsilon_i}$, as suggested by the authors. They use $\alpha$ to control the sparsity, i.e. the number of predicted edges. \par

Recently \cite{defazio2012convex} proposed a Lovasz extension approach to approximate node degree by a convex function. The convex function is a reweighed $l_1$ with larger penalty applied to edges with relatively larger strength. It turns out that such kind of convex function prior does not work well when we just need to predict a few edges, as shown in the experiments. Further, both \cite{liu2011learning} and \cite{defazio2012convex}  consider only the global degree distribution instead of the degree of each node. \par

\cite{tan2014learning} proposes a method (denoted as Hub) specifically for a graph with a few hubs and applies a group lasso penalty. In particular, they decompose $X$ as $Z+V+V^T$, where $Z$ is a sparse symmetric matrix and $V$ is a matrix whose column are almost entirely zero or non-zero. Intuitively, $Z$ describes the relationship between non-hubs and $V$ that between hubs. They formulate the problem as follows.
\begin{align}
\min_{V, Z} \ \ & F(X) + \lambda_1 \|Z\|_1 + \lambda_2 \|V\| + \lambda_3 \sum_{j=1}^p \|V_j\|_2 \nonumber\\
& s.t. \ \ X = Z + V + V^T
\end{align}
An ADMM algorithm is used to solve this problem. In our test, we use $\lambda_3=0.01$ to yield the best performance. Besides,we set $\lambda=\lambda_1=\lambda_2$ to produce a graph with a desired level of sparsity. \par

Our method uses $2$ hyper-parameters: $\gamma$ and $\beta$. Meanwhile, $\gamma$ is the hyper parameter for the power-law distribution and $\lambda$ controls sparsity.

\section{Results}
We tested our method on two real gene expression datasets and two types of simulated networks: (1) a scale-free network generated by Barabasi-Albert (BA) model \cite{albert2002statistical} and (2) a network with a few hub nodes.
We generated the data for the simulated scale-free network by its corresponding Multivariate Gaussian distribution. We compared our method (denoted as "DNS", short for "Dynamic Node-Specific Prior") with graphical lasso ("Glasso") \cite{friedman2008sparse}, neighborhood selection ("NS") \cite{meinshausen2006high}, reweighted $l_1$ regularization ("RW") \cite{liu2011learning}, Lovasz extenion ("Lovasz") \cite{defazio2012convex} and a recent hub detection method ("Hub") \cite{tan2014learning}.

\begin{figure*}[!Ht]
\begin{center}
\centerline{\includegraphics[width=\textwidth]{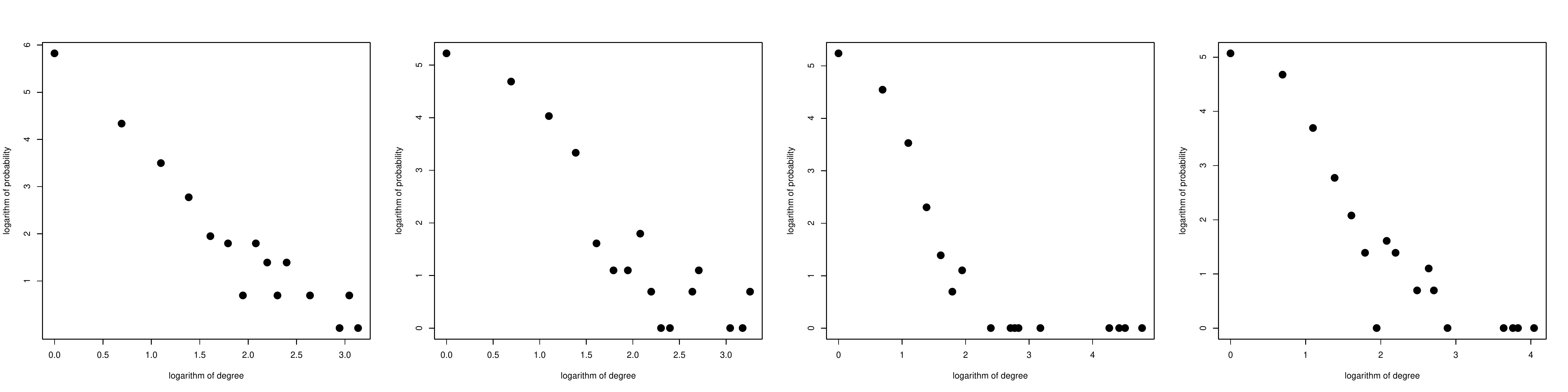}}
\caption{From left to right, the log-log degree distribution of (1) the true network and the estimated networks by (2) DNS (3) RW and (4) Glasso, respectively. Linear relationship is expected since the true network is scale-free. The network yielded by Glasso violates the power-law distribution most, as evidenced by a point close to (2,0). }
\label{scaledegree}
\end{center}
\end{figure*}

\subsection{Performance on Scale-Free Networks}
We generated a network of 500 nodes (p = 500) from the BA model. Each entry $\Omega_{uv}$ of the precision matrix is set to 0.3 if $(u,v)$ forms an edge, and $0$ otherwise. To make the precision matrix positive definite, we set the diagonal of $\Omega$ to the minimum eigenvalue of $\Omega$ plus 0.2. In total we generate 250 data samples from the corresponding multivariate Gaussian distribution $(i.e., n=250)$. The hyper-parameters of all the methods are set as described in the last section. \par
\begin{figure}[ht]
\begin{center}
\centerline{\includegraphics[width=70mm]{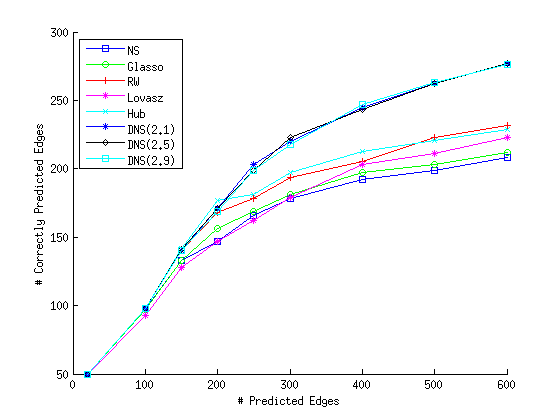}}
\caption{Simulation results on a scale free network. Gaussian Graphical Model is used with $n=250$ and $p=500$. X-axis is the number of predicted edges and Y-axis is the number of correctly predicted edges.}
\label{scaleacc}
\end{center}
\vskip -0.2in
\end{figure}
Our method is not sensitive to the hyper-parameter $\gamma$. As shown in Figure \ref{scaleacc}, a few different $\gamma$ values (2.1, 2.5, and 2.9) yield almost the same result. Hence we use $\gamma=2.5$ in the following experiments. \par

Moreover, as shown in Figure \ref{scaleacc}, our method DNS outperforms the others in terms of prediction accuracy. It is not surprising that both RW and Hub outperform Glasso and NS since the former two methods are specifically designed for scale-free or hub networks. Lovasz, which is also designed for scale-free networks, would outperform Glasso as the number of predicted edges increase.

Figure \ref{scaledegree} displays the log-log degree distribution of the true network and the networks estimated by DNS, RW and Glasso. Both DNS and RW yield networks satisfying the power-law distribution while Glasso does not, which confirms that both DNS and RW indeed favor scale-free networks.

\begin{figure}[ht]
\begin{center}
\centerline{\includegraphics[width=90mm]{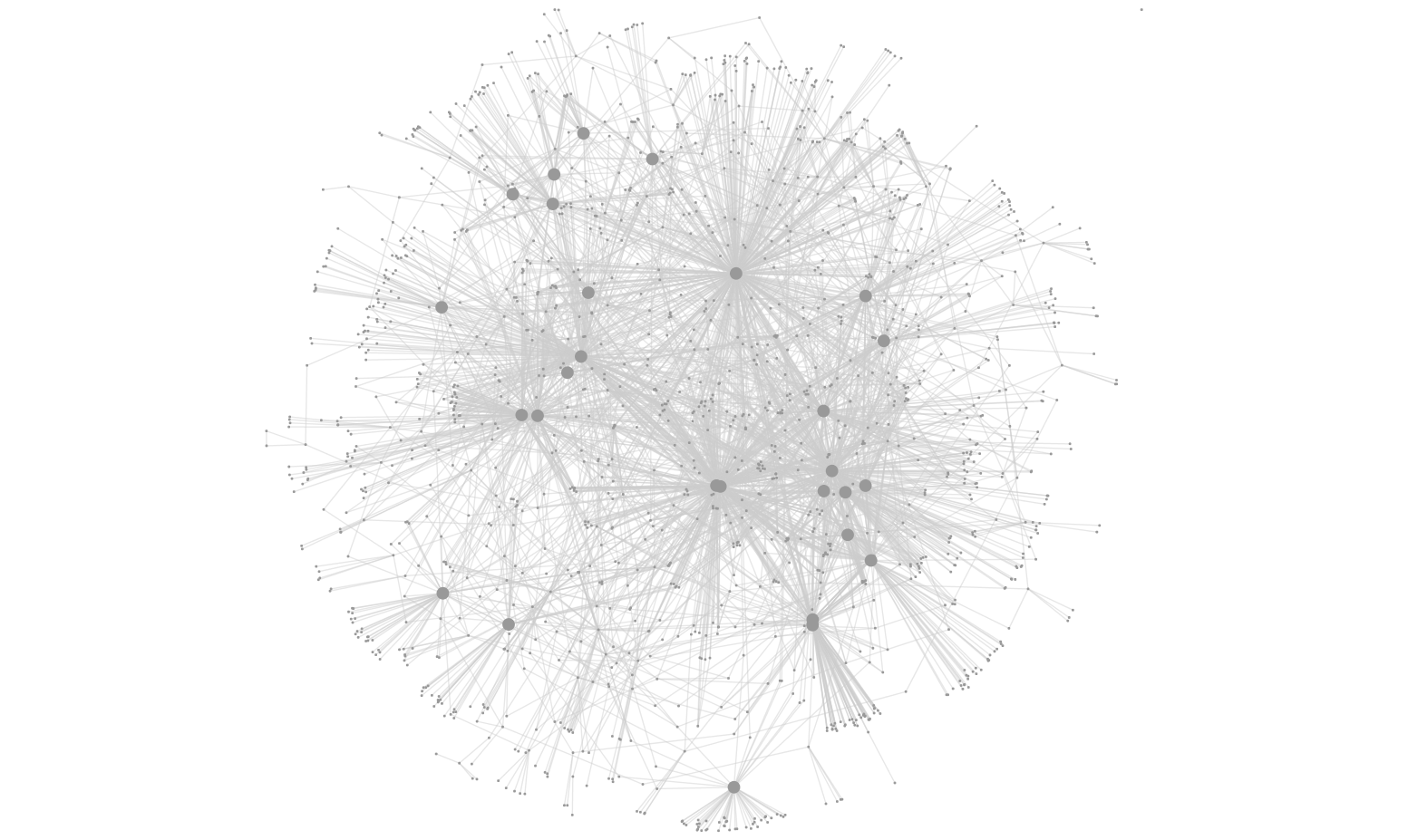}}
\caption{Visualization of the hub network with 1643 nodes and 3996 edges. For visualization purpose we ignore a connected component with less than or equal to 4 nodes. We also highlight the hub nodes, whose degree is at least 30.}
\label{truehub}
\end{center}
\vskip -0.2in
\end{figure}

\begin{figure}[ht]
\begin{center}
\centerline{\includegraphics[width=70mm]{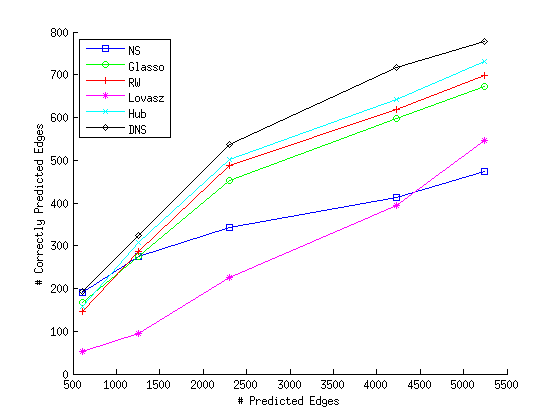}}
\caption{Simulation results of a hub network. Gaussian Graphical Model is used with $n=806$ and $p=1643$. X-axis is the number of predicted edges and Y-axis is the number of correctly predicted edges.}
\label{hubacc}
\end{center}
\vskip -0.2in
\end{figure}

\subsection{Performance on Hub Networks}
We also tested our method on a hub network, which contains a few nodes with very large degrees but not strictly follows the scale-free property. See Figure \ref{truehub} for a visualization, where larger dots indicate the hub nodes. Here we use the DREAM5 Network Inference Challenge dataset 1, which is a simulated gene expression data with 806 samples. DREAM5 also provides a ground truth network for this dataset. See \cite{marbach2012wisdom} for more details. \par
The result in Figure \ref{hubacc} shows that our method outperforms all the others, although our method is not specifically designed for hub networks. This shows that DNS also performs well in a graph with non-uniform degree distribution but without strict scale-free property.
\begin{figure}[ht]
\begin{center}
\centerline{\includegraphics[width=70mm]{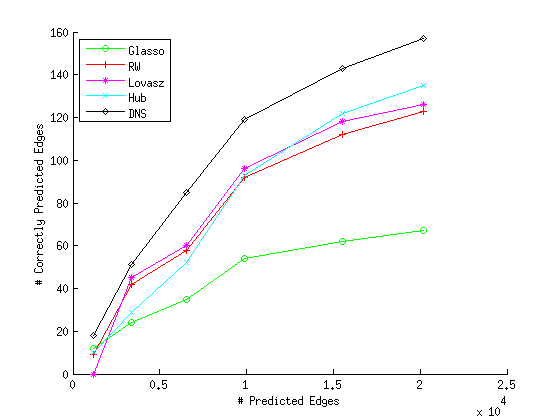}}
\caption{The performance of all the tested methods on a real gene expression data set (DREAM5 dataset 3). $X$-axis is the number of predicted edges and $Y$-axis is the number of correctly predicted edges.}
\label{realacc}
\end{center}
\end{figure}

\begin{figure*}[!Ht]
\begin{center}
\centerline{\includegraphics[width=\textwidth]{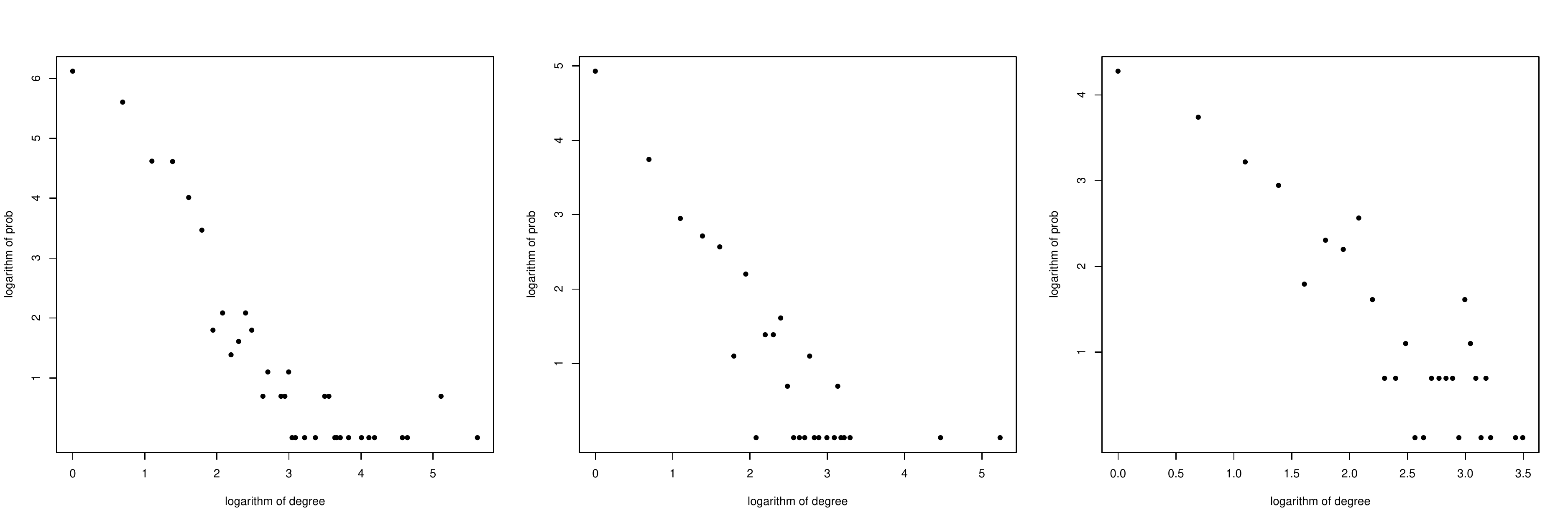}}
\caption{From left to right, the log-log degree distribution plot for (1) the true network and the estimated networks by (2) Glasso and (3) DNS, respectively.}
\label{realdeg}
\end{center}
\end{figure*}

\subsection{Gene Expression Data}
To further test our method, we used DREAM5 dataset 3 and 4 respectively. Dataset3 contains 805 samples and 4511 genes and its ground truth consists of 3902 edges. Dataset4 contains 536 samples and 5950 genes and its ground truth consists of 3940 edges. The two datasets are very challenging as the data is noisy and without Gaussian and scale-free property. For each dataset, up to $100,000$ predicted edges are allowed for submission for each team competing in the contest. See \cite{marbach2012wisdom} for a detailed description of the two data sets. We determine the hyper parameters of all the methods such that they output exactly the same number of edges. As shown in Figure \ref{realacc}, our method obtains much higher accuracy than the others on DREAM5 dataset 3. To compare the degree distribution of different methods, we chose the values of the hyper-parameters such that each method yields 4000 edges. The degree distributions of the resultant networks are shown in Figure \ref{realdeg}. As shown in this figure, our estimated network is more similar to a scale-free network. \par 
We also tested our result on DREAM5 dataset 4. As most algorithms are time consuming, we just run our method DNS and Glasso. According to our test, both our algorithm and Glasso perform not very good on this dataset (Actually, all DREAM5 competitors do not do well on this dataset \cite{marbach2012wisdom}). But our algorithm still outperforms Glasso in terms of accuracy. Actually, the accuracy of DNS is about two times of Glassso (e.g. when predicting about 19000 edges, Glasso generate 9 correct edges while DNS find 18 correct edges).

\section{Conclusions}
We have presented a novel node-specific degree prior to study the inference of scale-free networks, which are widely used to model social and biological networks.
Our prior not only promotes a desirable global node degree distribution,
but also takes into consideration the estimated degree of each individual node and the relative strength of all the possible edges adjacent to the same node.
To fulfill this, we have developed a ranking-based algorithm to dynamically model the degree distribution of a given network. The optimization problem resulting from our prior is quite challenging. We have developed a novel ADMM algorithm to solve it.\par

We have demonstrated the superior performance of our prior using simulated data and three DREAM5 datasets. Our prior greatly outperforms the others in terms of the number of correctly predicted edges, especially on the real gene expression data.\par

The idea presented in this paper can potentially be useful to other degree-constrained network inference problem. In particular, it might be applied to infer protein residue-residue interaction network from a multiple sequence alignment, for which we may predict the degree distribution of each residue using a supervised machine learning method.

\bibliography{scale_free}
\bibliographystyle{icml2015}

\end{document}


\maketitle 

This document is supplementary material for the paper entitled "Learning Scale-free Network by Dynamic Node Specific Prior" accepted by ICML 2015. In this document, when we refer to one equation, theorem or conclusion in the main text, we would explicitly mention it. Otherwise, we refer to an equation, theorem or conclusion in this document.

\section{Proof of equation $(8)$ in main text}

Equation (8) in the main text is equivalent to the claim that "The degree prior (7) in the main text favors a $p$ variable graph satisfying the given degree distribution $\{d_1,d_2,...d_p\}$". Now we prove \textbf{Equation 8} in the main text.
\begin{proof}
Assume the degree distribution of $E^{(2)}$ in \textbf{Equation 8} is $\{d_1',d_2',...,d_p'\}$ and inconsistent with the degree distribution $\{d_1,d_2,...d_p\}$. Without loss of generality, assume that $d_1'>d_1$ and $d_2'<d_2$. We can construct a graph $E'$ by moving one edge of variable $1$ to variable $2$. Actually we have

\begin{equation}
\sum_{u=1}^p{\frac{H\circ E_u'}{d_u}}-\sum_{u=1}^p{\frac{H\circ E^{(2)}_u}{d_u}} 
=\sum_{u=1}^2{\frac{H\circ E_u'}{d_u}}-\sum_{u=1}^2{\frac{H\circ E^{(2)}_u}{d_u}} 
=-\frac{H_{d_1'}}{H_{d_1}}+\frac{H_{d_2'+1}}{H_{d_2}}<0
\end{equation}

We can repeat such a construction process as long as $E'$ violates the degree distribution $\{d_1,d_2,...d_p\}$. Thus we have proved that \textbf{Equation $(8)$} in the main text holds. In other words, degree prior $(7)$ in the main text favors graphs following the given degree distribution.
\end{proof}

\section{Proof of Theorem 2}

\begin{proof}
Let $Y^{t+1}$ denote the output of \textbf{Algorithm 1}. It satisfies $[Y^{t+1},H,\delta^t] = [Y^{t},H,\delta^t]= [Y^{t+1},h].$
So 
\begin{align}
\arg\min_Y{\frac{1}{2}||Y-A||_F^2 + \lambda\sum_{i=1}^p{g(v)Y_{[v,Y^{t+1},h]}\circ H}}
\label{theorem1_condition}
\end{align}
is equivalent to 
\begin{align}
\arg\min_Y{\frac{1}{2}||Y-A||_F^2+\lambda\sum_{i=1}^p{g(v)Y_{[v,Y^{t+1},H,\delta^t]}\circ H}}
\label{last_step}
\end{align}

Since $[Y^{t+1},H,\delta^t] = [Y^{t},H,\delta^t]$, both $\delta$ and the objective function would not change, which means $Y^{t+1}$, the output of \textbf{Algorithm 1} in the main text, is also the solution of (\ref{last_step}) and (\ref{theorem1_condition}). Thus \textbf{Theorem 2} in the main text is proved.

\end{proof}
 
\section{Proof of Equation 21 in main text}
In this section, we prove Eq. (21) in the main text. We also list necessary definitions and lemmas for the next section, where we prove the correctness of our algorithm. \par
Actually, each subproblem in Eq. (20) in the main text can be written as

\begin{align}
\frac{1}{2}||Y_i-M_i||_F^2 + \sum_{k=1}^{p-1}{|Y_{i,(k)}|g(k)}
\label{objective}
\end{align}

Here, $\{g(1),g(2),...,g(p-1)\}$ is a sequence of positives. Given a feasible solution $Y_{i}$, we can assume $Y_{ii}=M_{ii}$, otherwise $Y_i$ cannot be the optimal solution. When we sort the elements in $Y_i$ and $M_i$, we do not consider their $i^{th}$ elements $Y_{i,i}$ and $M_{i,i}$. That is, we sort all the elements in $Y_i$ and $M_i$ excluding $Y_{i,i}$ and $M_{i,i}$.

\begin{figure}[h]
\centering
\includegraphics[width=0.6\textwidth]{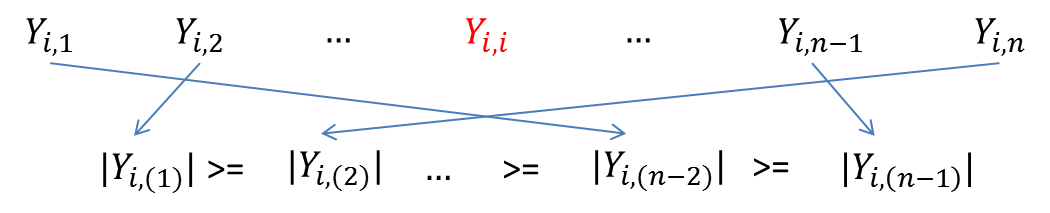}
\caption{Rank solution. The one-one correspondence is defined.}
\label{rank_solution}
\end{figure}

As shown in Figure \ref{rank_solution},
we may sort all the elements of $Y_{i}$ by their absolute values and obtain $|Y_{i,(1)}| \geq |Y_{i,(2)}| \geq ... \geq |Y_{i,(p-1)}|$. Obviously, there is one-one correspondence between $Y_{i,*}$ and $Y_{i,(*)}$. If $Y_{i,j}$ corresponds to $Y_{i,(k)}$, then we denote the correspondence as $Map(Y_{i,(k)})=j$. \\

Assume that $|Y_{i,(1)}|=|Y_{i,(2)}|=...=|Y_{i,(t)}|>|Y_{i,(t+1)}|$, we denote $\{Y_{i,(1)},Y_{i,(2)},...,Y_{i,(t)}\}$ as $C|_{Y_{i}}(1)$. Similarly, we can construct $C|_{Y_{i}}(2)$, $C|_{Y_{i}}(3)$ and so on. Assume $\cup_{k=1}^{p-1}Y_{i,(k)}$ can be clustered into $T(Y_i)$ groups, that is $\{C|_{Y_{i}}(1),C|_{Y_{i}}(2),...,C|_{Y_{i}}(T(Y_i))\}$. We use $|C|_{Y_{i}}(k)|$ to denote the number of elements in $C_{Y_{i}}(k)$, and for each element of $C|_{Y_{i}}(k)$, the absolute value should equal to $y|_{Y_{i}}(k)$. When context is clear, we use $C(k)$, $\hat{C}(k)$, $C'(k)$ and $C^{(l)}(k)$ for $Y_i$, $\hat{Y}_i$, $Y'_i$ and $Y^{(l)}_i$, respectively, and use $y(k)$, $T$, $\hat{y}(k)$, $\hat{T}$, $y'(k)$, $T'$, $y^{(l)}(k)$ and $T^{(l)}$ accordingly.

\begin{mydefinition}
Given a feasible solution $Y_i$ of (\ref{objective}), $Y_i$ is a stable solution of (\ref{objective}) if and only if for each $1\leq j \leq T$,

\begin{equation}
y(j)=max\{0,\frac{1}{|C(j)|}\sum_{s\in C(j) } {[|M_{i,Map(Y_{i,(s)})}|-g(s)]}\}
\end{equation}
\label{stable_def}
\end{mydefinition}

Given \textbf{Definition} (\ref{stable_def}), we have the following two lemmas. \par
\begin{mylemma}
The optimal solution $Y_i^{*}$ of (\ref{objective}) is a stable solution.
\label{min_condition}
\end{mylemma}

\begin{proof}
First we rewrite the objective function (\ref{objective}) as follows.
\begin{equation}
\frac{1}{2}\sum_{j=1}^p{(|Y_{i,j}|-|M_{i,j}|)^2}+\sum_{k=1}^{p-1}{|Y_{i,(k)}|g(k)}
\label{re_write_obj}
\end{equation}

Given a $j$ such that $1 \leq j \leq T(Y_i)$, if $y(j) \neq 0$, we can calculate the gradient over the absolute value of elements in $C(k)$, and set it to $0$, that is,

\begin{equation}
\sum_{s\in C(j)} {\{|Y_{i,(s)}|-|M_{i,Map(Y_{i,(s)})}|+g(s)\}} = 0
\end{equation}

Solving the above equation yields the following equation,
\begin{equation}
y(j)=\frac{1}{|C(j)|}\sum_{s\in C(j)} {[|M_{i,Map(Y_{i,(s)})}|-g(s)]}
\end{equation}

If $y(k)=0$, by calculating sub-gradient, it is easy to show that $\sum_{s\in C(j)} {[|M_{i,Map(Y_{i,(s)})}|-g(s)]} \leq 0$. Thus \textbf{Lemma} \ref{min_condition} is proved. 

\end{proof}

\begin{mylemma}
Solving (\ref{re_write_obj}) is equivalent to the following optimization problem,
\begin{equation}
Max_{\hat{Y}_i}{\sum_{k=1}^{p-1}{\hat{Y}_{i,(k)}^2}}
\end{equation}
subject to the condition that $\hat{Y}_i$ is a stable solution of (\ref{objective}) and (\ref{re_write_obj}).
\label{important_equi}
\end{mylemma}

\begin{proof}
According to \textbf{Lemma} \ref{min_condition}, the optimal solution is stable solution, thus we just need to consider the set of all stable solutions when solving (\ref{objective}). Given a stable solution $\hat{Y}_i$, Substitute $\hat{Y}_i$ to (\ref{objective}) and exclude $Y_ii$, we have following form,
\begin{align}
&\frac{1}{2}||\hat{Y}_i-M_i||_F^2 + \sum_{k=1}^{p-1}{|\hat{Y}_{i,(k)}|g(k)} \nonumber \\
=&\frac{1}{2}\sum_{k=1}^{p-1}{(|\hat{Y}_{i,(k)}|-|M_{i,(j)}|)^2} + \sum_{k=1}^{p-1}{|\hat{Y}_{i,(k)}|g(k)} \nonumber \\
=&\frac{1}{2}\sum_{k=1}^{p-1}{|\hat{Y}_{i,(k)}|^2}+\frac{1}{2}\sum_{k=1}^{p-1}{|M_{i,(k)}|^2}-\sum_{k=1}^{p-1}{|\hat{Y}_{i,(k)}|(|M_{i,(k)}|-g(k))} \nonumber \\
=&\frac{1}{2}\sum_{k=1}^{p-1}{|\hat{Y}_{i,(k)}|^2}+\frac{1}{2}\sum_{k=1}^{p-1}{|M_{i,(k)}|^2}-\sum_{j=1}^{\hat{T}}{|\hat{C}(j)|\hat{y}(j)^2} \nonumber \\
=&\frac{1}{2}\sum_{k=1}^{p-1}{|\hat{Y}_{i,(k)}|^2}+\frac{1}{2}\sum_{k=1}^{p-1}{|M_{i,(k)}|^2}-\sum_{k=1}^{p-1}{|\hat{Y}_{i,(k)}|^2}\nonumber \\
=&\frac{1}{2}[\sum_{s=1}^{p-1}{M_{i,(s)}^2} - \sum_{k=1}^{p-1}{|\hat{Y}_{i,(k)}|^2}]
\end{align}

As $\sum_{s=1}^{p-1}{M_{i,(s)}^2}$ is constant, so it is reasonable to conclude that solving (\ref{objective}) is equivalent to maximize $\sum_{k=1}^{p-1}{|\hat{Y}_{i,(k)}|^2}$ in the space of all stable solutions. Thus we conclude that \textbf{Lemma 2} is correct.
\end{proof}

Based on \textbf{Lemma 2}, we can define the partial order relationship among stable solutions.

\begin{mydefinition}
Given two stable solutions $Y_i^{(1)}$ and $Y_i^{(2)}$ of (\ref{objective}), we say $Y_i^{(1)} \succ Y_i^{(2)}$ if and only if the following inequality holds
\begin{equation}
\sum_{j=1}^{p-1}{{Y^{(1)}_{i,(j)}}^2}>\sum_{j=1}^{p-1}{{Y^{(2)}_{i,(j)}}^2}
\end{equation}
\end{mydefinition}

Let $Y_i^{*}$ be the optimal solution. Then there is no feasible solution $Y_i$ such that $Y_i \succ Y_i^{*}$. Before proving \textbf{Equation 21} in the main text, we show the following lemma.

\begin{mylemma}
Let $\{x_1,x_2,...,x_n\}$ be a non-negative non-decreasing sequence and  $\{\delta_1,\delta_2,...,\delta_n\}$ be a set of non-negative numbers satisfying $\sum_{i=1}^s{\delta_i} \geq \sum_{i=s+1}^{n}{\delta_i}>0$. Then the following inequality holds,
\begin{equation}
\sum_{i=1}^s{(x_i+\delta_i)^2} + \sum_{i=s+1}^n{(x_i-\delta_i)^2} \geq \sum_{i=1}^n{{x_i}^2}
\end{equation}
\label{trick}
\end{mylemma}

\begin{proof}
\begin{eqnarray}
\sum_{i=1}^s{(x_i+\delta_i)^2} + \sum_{i=s+1}^n{(x_i-\delta_i)^2} & = & \sum_{i=1}^n{x_i^2}+\sum_{i=1}^n{\delta_i^2}+2[\sum_{i=1}^{s}{x_i\delta_i}-\sum_{i=s+1}^{n}{x_i\delta_i}] \nonumber \\
 & \geq & \sum_{i=1}^n{x_i^2}+\sum_{i=1}^n{\delta_i^2}+ 2[\sum_{i=1}^{s}{x_s\delta_i}-\sum_{i=s+1}^{n}{x_{s+1}\delta_i}] \nonumber \\
& = & \sum_{i=1}^n{x_i^2}+\sum_{i=1}^n{\delta_i^2}+ 2[x_s\sum_{i=1}^s{\delta_i}-x_{s+1}\sum_{i=s+1}^{n}{\delta_i}] \nonumber \\
& \geq & \sum_{i=1}^n{x_i^2}+\sum_{i=1}^n{\delta_i^2}+ 2[(x_s-x_{s+1})]\sum_{i=s+1}^n{\delta_i} \nonumber \\
& \geq & \sum_{i=1}^n{x_i^2}+\sum_{i=1}^n{\delta_i^2} \nonumber \\
& \geq & \sum_{i=1}^n{x_i^2} \nonumber
\end{eqnarray}

Note that strict inequality holds as long as at least two elements in $\{x_1,x_2,...,x_n\}$ are different.
\end{proof}

Actually, proving \textbf{Equation 21} is equivalent to proving \textbf{Lemma} \ref{equal_lemma}.
\begin{mylemma}
Given the optimal solution $Y_{i}^{*}$, for every $j$ such that $1 \leq j \leq T^{*}$, the following two sets are equal,
\begin{equation}
\{M_{i,Map(Y_{i,(\sum_{x=1}^{j-1}{|C^{*}(x)|}+1)})}, M_{i,Map(Y_{i,(\sum_{x=1}^{j-1}{|C^{*}(x)|}+2)})}, ..., M_{i,Map(Y_{i,(\sum_{x=1}^{j}{|C^{*}(x)|})})}\} \nonumber
\end{equation}

\begin{equation}
\{M_{i,(\sum_{x=1}^{j-1}{|C^{*}(x)|}+1)}, M_{i,(\sum_{x=1}^{j-1}{|C^{*}(x)|}+2)}, ..., M_{i,(\sum_{x=1}^{j}{|C^{*}(x)|})}\} \nonumber
\end{equation}
\label{equal_lemma}
\end{mylemma}

Further, to prove \textbf{Lemma} \ref{equal_lemma}, we just need to prove the following \textbf{Lemma} \ref{easier_lemma}, which is much easier to prove.

\begin{mylemma}
Let $Y^{(0)}_i$ be a stable solution. If there exists a number $q^{(0)}$ ($1 \leq q^{(0)} \leq T^{(0)}$) and two numbers $s_1,s_2$ where $\sum_{x=1}^{q^{(0)}-1}{|C(x)|}<s_1 \leq \sum_{x=1}^{q^{(0)}}{|C(x)|}$ and $\sum_{x=1}^{q^{(0)}}{|C(x)|}<s_2 \leq \sum_{x=1}^{q^{(0)}+1}{|C(x)|}$, such that $M_{i,Map^{(0)}(Y_{i,(s_1)})}<M_{i,Map^{(0)}(Y_{i,(s_2)})}$, then there is another stable solution $Y_i'$, such that $Y_i' \succ Y_i$. Here we use $Map^{(t)}$ to denote the mapping of $Y^{(t)}_i$.

\label{easier_lemma}
\end{mylemma}

\begin{proof}
To prove \textbf{Lemma} \ref{easier_lemma}, we just need to show that we can construct  a $Y_i'$ described in \textbf{Lemma} \ref{easier_lemma}. \par
We first construct $Y^{(1)}_i$ as follows. We set $Map^{(1)}(Y_{i,(s_1)})=Map^{(0)}(Y_{i,(s_2)})$ and $Map^{(1)}(Y_{i,(s_2)})=Map^{(0)}(Y_{i,(s_1)})$ and then set each component of $Y^{(1)}_i$ as follows,
\begin{eqnarray}
Y^{(1)}_{i,i} &=& Y^{(0)}_{i,i} \nonumber \\
Y^{(1)}_{i,(s)} &=& Y^{(0)}_{i,(s)}\quad if \quad 1 \leq s \leq \sum_{x=1}^{q^{(0)}-1}{|C^{(0)}(x)|} \quad or \quad \sum_{x=1}^{q^{(0)}+1}{|C^{(0)}(x)|} < s \leq n-1 \nonumber \\
Y^{(1)}_{i,(s)} &=& \frac{1}{|C^{(0)}(q^{(0)})|}\sum_{j\in C(q^{(0)}) }{[|M_{i,Map^{(1)}(Y_{i,(j)})}|-g(j)]} \quad if \quad \sum_{x=1}^{q^{(0)}-1}{|C(x)|} < s \leq \sum_{x=1}^{q^{(0)}}{|C(x)|} \nonumber \\
Y^{(1)}_{i,(s)} &=& \frac{1}{|C^{(0)}(q^{(0)}+1)|}\sum_{j\in C(q^{(0)}+1)} {[|M_{i,Map^{(1)}(Y_{i,(j)})}|-g(j)]} \quad if \quad \sum_{x=1}^{q^{(0)}}{|C(x)|} < s \leq \sum_{x=1}^{q^{(0)}+1}{|C(x)|} \nonumber
\end{eqnarray}

If $Y^{(1)}_i$ is consistent, i.e., satisfying Condition (\ref{consistent}), then $Y^{(1)}_i$ is a stable solution. In this case we can simply set $Y'_i=Y^{(1)}_i$.
\begin{equation}
y^{(1)}(1)>...>y^{(1)}(q^{(0)}-1)>y^{(1)}(q^{(0)})>y^{(1)}(q^{(0)}+1)>y^{(1)}(q^{(0)}+2)>...>y^{(1)}(T(y^{(1)}))
\label{consistent}
\end{equation}

Otherwise, let $q^{(1)}=q^{(0)}$ and $l^{(1)}=q^{(0)}+1$, and then continue the following process. \par
Assume that we have constructed $Y^{(k)}_i$. We say that $Y_i^{(k)}$ is "left-consistent" if
\begin{equation}
y^{(k)}(q^{(k)})<y^{(k)}(q^{(k)}-1).
\label{left_consis}
\end{equation}

And similarly, we say that $Y_i^{(k)}$ is "right-consistent" if
\begin{equation}
y^{(k)}(q^{(k)}+2)<y^{(k)}(q^{(k)}+1).
\label{right_consis}
\end{equation}

Then we start to construct $Y^{(k+1)}_i$. If $Y_i^{(k)}$ is not "left-consistent", set $q^{(k+1)}=q^{(k)}-1$, merge $C^{(k)}(q^{(k)})$ and $C^{(k)}(q^{(k)}-1)$ to one group $C^{(k+1)}(q^{(k+1)})$ and set $y^{(k+1)}(q^{(k+1)})$ as follows
\begin{equation}
y^{(k+1)}(q^{(k+1)})= \frac{|C^{(k)}(q^{(k)}-1)|y^{(k)}(q^{(k)}-1)+|C^{(k)}(q^{(k)})|y^{(k)}(q^{(k)})}{|C^{(k)}(q^{(k)}-1)|+|C^{(k)}(q^{(k)})|}
\end{equation}

Otherwise, we do not need merge the two groups, but just set $q^{(k+1)}=q^{(k)}$ and  $y^{(k+1)}(q^{(k+1)})=y^{(k)}(q^{(k)})$.  \par
Also, if $Y_i^{(k)}$ is not "right-consistent", then merge $C^{(k)}(q^{(k)}+1)$ and $C^{(k)}(q^{(k)}+2)$ into one group $C^{(k+1)}(q^{(k+1)}+1)$. Set $l^{(k+1)}=l^{(k)}+1$, and set $y^{(k+1)}(q^{(k+1)}+1)$ as follows,

\begin{equation}
y^{(k+1)}(q^{(k+1)}+1)= \frac{|C^{(k)}(q^{(k)}+1)|y^{(k)}(q^{(k)}+1)+|C^{(k)}(q^{(k)}+2)|y^{(k)}(q^{(k)}+2)}{|C^{(k)}(q^{(k)}+1)|+|C^{(k)}(q^{(k)}+2)|}
\end{equation}

Otherwise, set $l^{(k+1)}=l^{(k)}$ and $y^{(k+1)}(q^{(k+1)}+1)=y^{(k)}(q^{(k)}+1)$. We then construct the rest part of $Y^{(k+1)}_i$ as follows,

\begin{eqnarray}
Y^{(k+1)}_{i,i} &=& Y^{(k)}_{i,i} \nonumber \\
Y^{(k+1)}_{i,(s)} &=& Y^{(k)}_{i,(s)} \quad if \quad 1\leq s \leq \sum_{x=1}^{q^{(k+1)}-1}{|C^{(k)}(x)|} \nonumber \\
Y^{(k+1)}_{i,(s)} &=& Y^{(k)}_{i,(s)} \quad if \quad \sum_{x=1}^{q^{(k+1)}+1}{|C^{(k+1)}(x)|} < s <n \nonumber
\end{eqnarray}

Then we get a new solution $Y_i^{(k+1)}$. If $Y_i^{(k+1)}$ is consistent , then we denote $Y_i' = Y_i^{(k+1)}$. Otherwise, we continue the process described above to construct $Y_i^{(k+2)}$. Note that we would definitely stop as such kind of construction process will not go on forever. When the process stopped, we would definitely get a stable solution $Y_i'=Y_i^{(t)}$. According \textbf{Lemma} \ref{important_equi}, to prove that $Y_i^{(t)} \succ Y_i$ we just need to prove the following inequality holds,

\begin{equation}
\sum_{a=q^{(t)}}^{l^{(t)}}{|C^{(0)}(a)|{y^{(0)}(a)}^2}<|C^{(t)}(q^{(t)})|{y^{(t)}(q^{(t)})}^2+|C^{(t)}(q^{(t)}+1)|{y^{(t)}(q^{(t)}+1)}^2
\end{equation}

If $y^{(t)}(q^{(t)}+1) \neq 0$, we have the following equality holds,

\begin{equation}
\sum_{a=q^{(t)}}^{q^{(1)}}{Y^{(t)}_{i,a}-Y_{i,a}}=\sum_{a=l^{(1)}}^{l^{(t)}}{Y_{i,a}-Y^{(t)}_{i,a}}
\end{equation}

If $y^{(t)}(q^{(t)}+1)=0$, then we have

\begin{equation}
\sum_{a=p'}^{q^{(1)}}{Y'_{i,a}-Y_{i,a}}>\sum_{a=l^{(1)}}^{l'}{Y_{i,a}-Y'_{i,a}}
\end{equation}

According to \textbf{Lemma} \ref{trick} we have

\begin{equation}
\sum_{a=q^{(t)}}^{l^{(t)}}{|C^{(0)}(a)|{y^{(0)}(a)}^2}=\sum_{a=q^{(t)}}^{l^{(t)}}{Y^2_{i,a}}<\sum_{a=q^{(t)}}^{l^{(t)}}{{Y^{(t)}_{i,a}}^2}=|C^{(t)}(q^{(t)})|{y^{(t)}(q^{(t)})}^2+|C^{(t)}(q^{(t)}+1)|{y^{(t)}(q^{(t)}+1)}^2
\label{five_proved}
\end{equation}
According to (\ref{five_proved}), \textbf{Lemma} \ref{easier_lemma} is proved. Thus \textbf{Lemma} \ref{equal_lemma} and \textbf{Equation 21} in the main text are also proved.
\end{proof}

\section{Algorithm 2 in main text (Edge ranking updating)}

\subsection{Partition Problem}
Based on the conclusion of last section, solving (\ref{objective}) can be reduced to the following problem (\ref{sorted_problem})
\begin{equation}
\min\big (  \sum_{i=1}^{p-1}{|Y_{v,(i)}-M_{v,(i)}|_F^2}+\sum_{i=1}^{p-1}{g(i)|Y_{v,(i)}|} \big ).
\label{sorted_problem}
\end{equation}

\begin{figure}[h]
\centering
\includegraphics[width=0.6\textwidth]{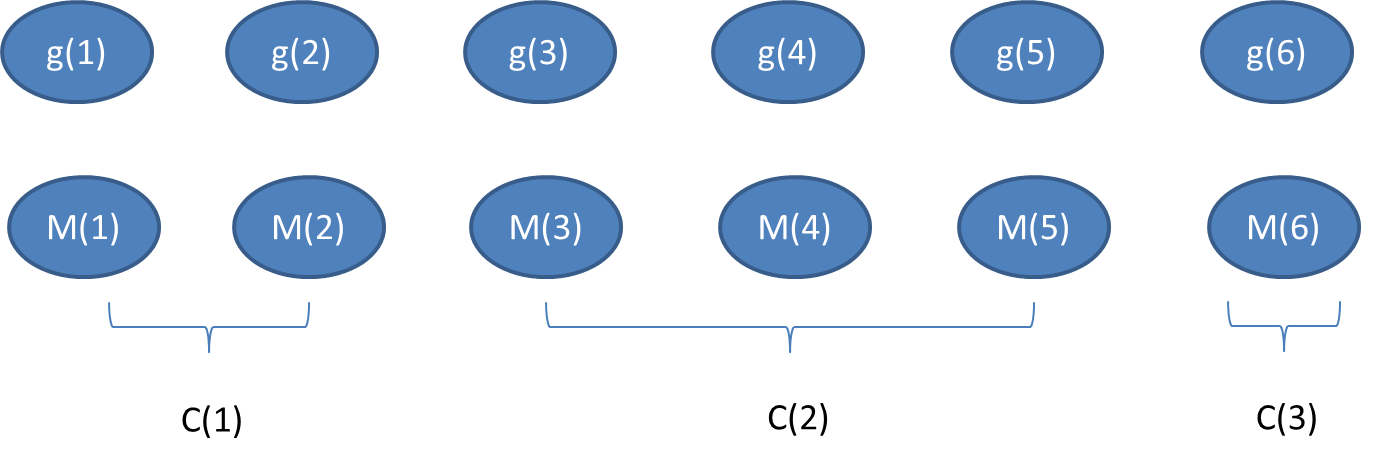}
\caption{Partition of ordered sequences.}
\label{equal_partition}
\end{figure}

This can be understood as partitioning ordered sequences ($M_{a,(1)}$, $M_{a,(2)}$,...,$M_{a,(p-1)}$) into optimal blocks, as shown in Figure \ref{equal_partition}, such that the square sum of the final solution is maximized, as claimed in \textbf{Lemma} \ref{important_equi}. In Figure \ref{equal_lemma}, we use $M(i)$ to denote $M_{a,(i)}$ for some $a$. 

\subsection{Property of optimal solution}
Given a number $t$ ($1\leq t < p$), let $Y_{v,1:t}^{*}$ denote the optimal solution of the following sub-problem
\begin{align}
SUB(t) = \min\big (  \sum_{i=1}^t{|Y_{v,(i)}-M_{v,(i)}|_F^2}+\sum_{i=1}^t{g(i)|Y_{v,(i)}|} \big )
\label{sub_problem}
\end{align}

Actually, the cluster structure $\{C|_{Y_{v,1:t}^{*}}(1), C|_{Y_{v,1:t}^{*}}(2),...,C|_{Y_{v,1:t}^{*}}(T(Y_{v,1:t}^{*}))\}$ of $Y_{v,1:t}^{*}$ corresponds to the optimal partition of ($M_{v,(1)}$, $M_{v,(2)}$,...,$M_{v,(t)}$). Now we show two properties of the cluster structure.
\begin{itemize}
\item[1] Denote $C|_{Y_{v,1:t}^{*}}(k)=\{Y_{v,(a)},Y_{v,(a+1)},...,Y_{v,(b)}\}$. For any $a\leq s < b$, we have $\frac{\sum_{i=a}^{s}{|M_{v,(i)}|-g(i)}}{s-a+1} \leq \frac{\sum_{i=s+1}^{b}{|M_{v,(i)}|-g(i)}}{b-s}$.
\item[2] For any $b<s<p$, we have $\frac{\sum_{i=a}^{b}{|M_{v,(i)}|-g(i)}}{b-a+1} > \frac{\sum_{i=b+1}^{s}{|M_{v,(i)}|-g(i)}}{s-b}$.
\end{itemize}

Here we show the proof of \textbf{Property 1} and \textbf{Property 2} can be proved  similarly.
\begin{proof}
We assume that \textbf{Property 1} does not hold, that there exists a $s$, such that $\frac{\sum_{i=a}^{s}{|M_{v,(i)}|-g(i)}}{s-a+1} > \frac{\sum_{i=s+1}^{b}{|M_{v,(i)}|-g(i)}}{b-s}$. \par
For simplicity, let $Y_{v,1:t}^{*}=\{C(1),C(2),...,C(T)\}$. We can construct $Y^{(1)}$ as follows.
\begin{align}
Y^{(1)}=\{ C(1), C(2),...,C(k-1),C_{left}(k),C_{right}(k),C(k+1),...,C(T)\}
\label{new_solution}
\end{align}
Each element of $C_{left}(k)$ equals to $\frac{\sum_{i=a}^{s}{|M_{v,(i)}|-g(i)}}{s-a+1}$ while each element of $C_{right}(k)$ equals to $\frac{\sum_{i=s+1}^{b}{|M_{v,(i)}|-g(i)}}{b-s}$. If (\ref{new_solution}) is consistent (i.e., satisfying (\ref{consistent})), then the construction process ends. Otherwise, we start to construct $Y^{(2)}$. The process is very similar to what we have done in proving \textbf{Lemma} \ref{easier_lemma}. If $Y^{(1)}$ is not left-consistent (see (\ref{left_consis})), then we combine $C(k-1)$ and $C_{left}(k)$ and if $Y^{(1)}$ is not right-consistent (see (\ref{right_consis})), then we combine $C_{right}(k)$ and $C(k+1)$. If $Y^{(2)}$ is not consistent, we will continue such a process until a consistent solution $Y^{(m)}$ is obtained, By using the same technique in proving \textbf{Lemma} \ref{easier_lemma}, we can show that $Y^{(m)} \succ Y_{v,1:t}^{*}$, which contradicts to the assumption that $Y_{v,1:t}^{*}$ is optimal.
\end{proof}
\subsection{Correctness of Theorem 3 in main text}
Based on the two properties, we now show that \textbf{Theorem 3} in the main text is correct. \par
\begin{proof}
For simplicity, we still denote $Y_{v,1:t}^{*}=\{C(1),C(2),...,C(T)\}$, and $C(T+1)=\{|M_{v,(t+1)}|-g(t+1)\}$. Assume that $k$ is the largest value we mentioned in \textbf{Theorem 3} in the main text. \par
For any $1\leq x \leq k$, $\{C(1),...,C(x-1),C_x\}$ would be a consistent solution of $SUB(t+1)$ (see definition \ref{sub_problem}), where $C_x$ is a set with $\Sigma_{s=x}^{T+1}{|C(s)}|$ elements whose absolute value all equal to $y_x$
\begin{align}
y_x = \max \{0,\frac{\sum_{i \in {\bigcup_{s=x}^{T+1}{C(s)}}}\{|M_{v,(i)}|-g(v)\}}{\Sigma_{s=x}^{T+1}{|C(s)}|} \}
\label{solution_form}
\end{align}
Obviously, among these solutions, $\{C(1),...,C(k-1),C_k\}$ is the largest one based on \textbf{Lemma} \ref{important_equi}. We now just need to prove that the optimal solution $Y_{v,1:t+1}^{*}$ of $SUB(t+1)$ is actually one of the $k$ solutions $\{C(1),...,C(x-1),C_x\}$ for $1\leq x \leq k$. \par
Assume $Y_{v,1:t+1}^{*}=\{C(1), C(2), ... , C(b-1),C'(b),C'(b+1),...,C'(T') \}$, which means the $Y_{v,1:t+1}^{*}$ and $Y_{v,1:t}^{*}$ share the first $b-1$ clusters, and $C'(b) \neq C(b)$. We would like to prove $b=T'$. If $b \neq T'$, there are two cases
\begin{itemize}
\item[1] $|C'(b)|>|C(b)|$. Denote $C'(b)=\{ Y_{v,(m)},Y_{v,(m+1)},...,Y_{v,(q)}\}$ while $C(b)=\{ Y_{v,(m)},Y_{v,(m+1)},...,Y_{v,(n)}\}$. According to \textbf{Property 2}, we have $\frac{\sum_{i=m}^{n}{|M_{v,(i)}|-g(i)}}{n-m+1} > \frac{\sum_{i=n+1}^{q}{|M_{v,(i)}|-g(i)}}{q-n}$; However, according to \textbf{Property 1}, we have $\frac{\sum_{i=m}^{n}{|M_{v,(i)}|-g(i)}}{n-m+1} \leq \frac{\sum_{i=n+1}^{q}{|M_{v,(i)}|-g(i)}}{q-n}$. Contradiction! Thus, the assumption $|C'(b)|>|C(b)|$ does not hold.
\item[2] $|C'(b)|<|C(b)|$. We would lead to similar contradiction! Thus, the assumption $|C'(b)|<|C(b)|$ does not hold.
\end{itemize}
Note that $C'(b) \neq C(b)$ is equivalent to $|C'(b)| \neq |C(b)|$. So the only possibility is the assumption that $b \neq T'$ is wrong. Thus $b$ should equal to $T'$, and this is equivalent to say that $Y_{v,1:t+1}^{*}$ is always among the $k$ solutions. \par
Proved.
\end{proof}

\subsection{$O(p\log(p))$ algorithm}
\textbf{Theorem 3} in the main text defines the problem structure suitable for dynamic programming. We further notice the following lemma, which is not hard to prove, leads to the $O(p\log(p))$ algorithm in the main text.
\begin{mylemma}
For simplicity, we denote $Y_{v,1:t}^{*}=\{C(1),C(2),...,C(T)\}$ and $C(T+1)=\{|M_{v,(t+1)}|-g(t+1)\}$. 
Given $\{C(1),...,C(x-1),C_x\}$, where $C_x$ is a set with $\Sigma_{s=x}^{T+1}{|C(s)}|$ elements having the same value as $y_x$
\begin{align}
y_x = \max \{0,\frac{\sum_{i \in {\bigcup_{s=x}^{T}{C(s)}}}\{|M_{v,(i)}|-g(v)\}}{\Sigma_{s=x}^{T}{|C(s)}|} \}
\end{align}
If $y(x-1)<y_x$, then for any $s$ where $x\leq s \leq T+1$, $\{C(1),...,C(s-1),C_s\}$ is not consistent as $y(s-1)<y_s$. \par 
Similarly, if $y(x-1)>y_x$, then for any $s$ where $1\leq s \leq x$, $\{C(1),...,C(s-1),C_s\}$ is consistent as $y(s-1)>y_s$. \par 
\label{binary_lemma}
\end{mylemma}
\textbf{Lemma} \ref{binary_lemma} allows us to check the largest $k$ value in $O(\log(p))$, which is the time complexity to build solution of $SUB(t+1)$ from $Y_{v,1:t}^{*}$. Thus, we can construct a $Y^{*}_v=Y^{*}_{v,1:p-1}$ with time complexity $O(p\log(p))$.